\documentclass[runningheads]{llncs}

\usepackage{eccv}

\usepackage{eccvabbrv}

\usepackage{graphicx}
\usepackage{multicol,multirow,adjustbox}
\usepackage{booktabs}
\usepackage{soul}
\usepackage{makecell}
\usepackage{wrapfig}
\usepackage[dvipsnames,table]{xcolor}
\usepackage[accsupp]{axessibility}  % Improves PDF readability for those with disabilities.

\usepackage{hyperref}

\usepackage{orcidlink}

\definecolor{mygreen}{RGB}{0,128,0}
\definecolor{mygray}{gray}{0.9}

\begin{document}

% ---------------------------------------------------------------
% TODO REVIEW: Replace with your title
\title{\textit{\textmu}Flow: Leveraging Average Images for Improving Generalisation of Deepfake Faces Detectors}

% TODO REVIEW: If the paper title is too long for the running head, you can set
% an abbreviated paper title here. If not, comment out.
\titlerunning{\textit{\textmu}Flow}

% TODO FINAL: Replace with your author list. 
% Include the authors' OCRID for the camera-ready version, if at all possible.
\author{Orazio Pontorno\thanks{These authors equally contributed to this work.}\inst{1}\orcidlink{0009-0009-0381-9971} \and
Mattia Litrico$^*$\inst{1}\orcidlink{0000-0002-4914-174X} \and
Luca Guarnera\inst{1}\orcidlink{0000-0001-8315-351X} \and Mario Valerio Giuffrida\inst{2}\orcidlink{0000-0002-5232-677X} \and Sebastiano Battiato \inst{1}\orcidlink{0000-0001-6127-2470}}

% TODO FINAL: Replace with an abbreviated list of authors.
\authorrunning{O. Pontorno et al.}

\institute{
University of Catania, Catania, Italy
\and
University of Nottingham, Nottingham, UK
\email{
\{orazio.pontorno,mattia.litrico\}@phd.unict.it,
\{luca.guarnera,sebastiano.battiato\}@unict.it,
valerio.giuffrida@nottingham.ac.uk
}
}
\maketitle

\begin{abstract}
Current generative models, including GANs and diffusion models, have reached an outstanding level of photorealism, posing significant risks to privacy and security. To ensure real-world applicability, deepfake detectors must generalise effectively to unseen generators. However, most existing approaches rely on supervised training with both real and fake images, which limits their generalisation especially across generators categories (e.g. GANs vs DMs). In this work, we introduce \textbf{\textit{\textit{\textmu}Flow}}, a one-class deepfake detector trained only on real images without relying on pseudo-deepfakes or synthetic artifacts. Our approach builds on the observation that averaging multiple images amplifies consistent generative traces, producing highly discriminative feature representations. We leverage this property by modelling the distribution of features extracted from averaged images and training a normalizing flow to align the feature space of individual images with this distribution. This alignment yields a likelihood-based criterion that separates real and fake samples while promoting strong generalisation.
We evaluate \textbf{\textit{\textit{\textmu}Flow}} on a fully out-of-distribution setting, where both real and fake datasets are unseen during training. Experimental results show that our method significantly outperforms SOTA detectors. Project page: \href{https://opontorno.github.io/MuFlow/}{opontorno.github.io/MuFlow}.
\end{abstract}

\section{Introduction}
\label{sec:intro}

The generation of deepfakes has reached a worrisome level of photorealism, increasingly raising privacy concerns among the general public. Deepfakes are obtained by either forging a real image \cite{man1,man2,man3,man4}, or creating them from scratch using generative models \cite{gen1,gen2,gen3,gen4}, such as GANs \cite{goodfellow2014generative} or diffusion models (DMs) \cite{ho2020denoising}.  Given the recent strides in generative models, detecting forgeries has become increasingly challenging using conventional methodologies. Nonetheless, these models introduce distinctive traces that can be detected by deep learning techniques \cite{yang2025d, pontorno2024exploitation}. 
Most of the current deepfake detectors are trained using a labelled training set comprising both real and fake images \cite{soudy2024deepfake,ke2023df,zhang2019detecting,wang2021fakespotter}.\\
\begin{wrapfigure}{r}{0.5\textwidth}
    \centering
    \includegraphics[width=\linewidth]{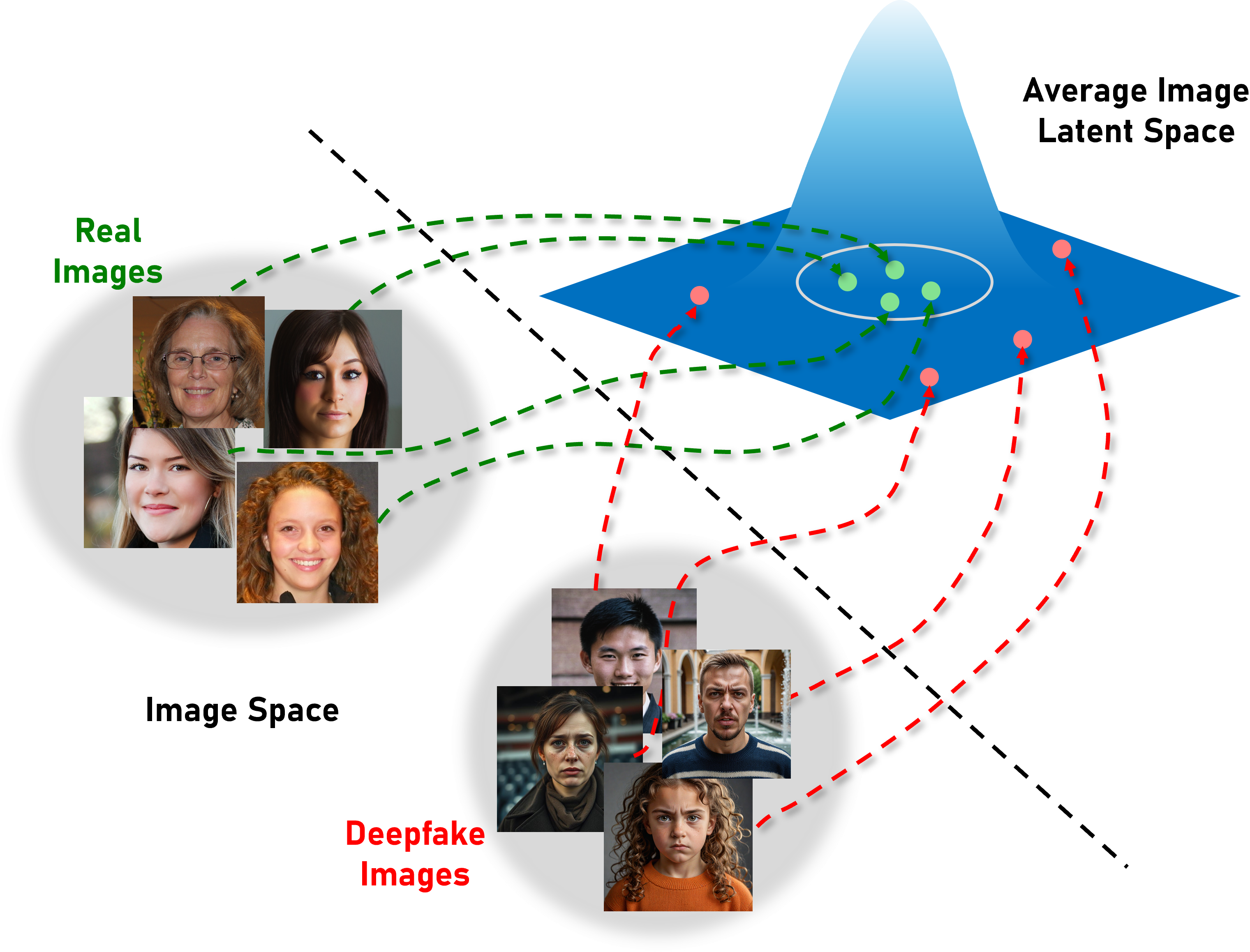}
    \caption{\textbf{\textit{\textmu}Flow} overview: image features are mapped into a discriminative distribution trained from average image features determined from just real images. In this way, fake images obtain a lower likelihood at test time, as being considered out-of-distribution samples.}
    \label{fig:visual_abs}
\end{wrapfigure}
These approaches can easily detect deepfakes from generators seen during training (in-distribution) but struggle to generalise to unseen generators (out-of-distribution).
 
 For this reason, recent approaches have focused on assessing the generalisation capability of deepfake detectors. In fact, recent works focused on improving robustness through searching for universal traces \cite{tan2024frequency,tan2024rethinking,yang2025d}, vision-language models (VLMs) \cite{tan2025c2p,CLIP2} or pretrained models \cite{UFD,SVM}.
However, these approaches are trained on both real and fake images, and tend to overfit to distinctive traces injected by specific generators seen during training \cite{yang2025d}. This limits their generalisation performance, as generative pipelines improve much faster than detector datasets can be updated.

To overcome this issue, some approaches proposed to leverage real-only training \cite{diff_fake,larue2023seeable,UNTAG,yuan2025clip}. Such approaches offer a promising direction to improve the generalisability of deepfake detectors since they are not constrained by specific generators. However, several approaches \cite{larue2023seeable,UNTAG,yuan2025clip} rely on generating pseudo-deepfakes for learning a discriminative features space~\cite{torralba2003statistics}, limiting their performance to the quality of such pseudo-deepfake.
A different approach is OC-FakeDect \cite{khalid2020oc} that maps real samples into a normal Gaussian distribution. However, this approach is characterized by limited discriminative power.
Therefore, there is a lack of deepfake detectors able to generalise to unseen generators, without requiring the need of fake samples during training but still maintaining high discriminability.

\begin{figure}[t]
    \centering
    \includegraphics[width=\linewidth]{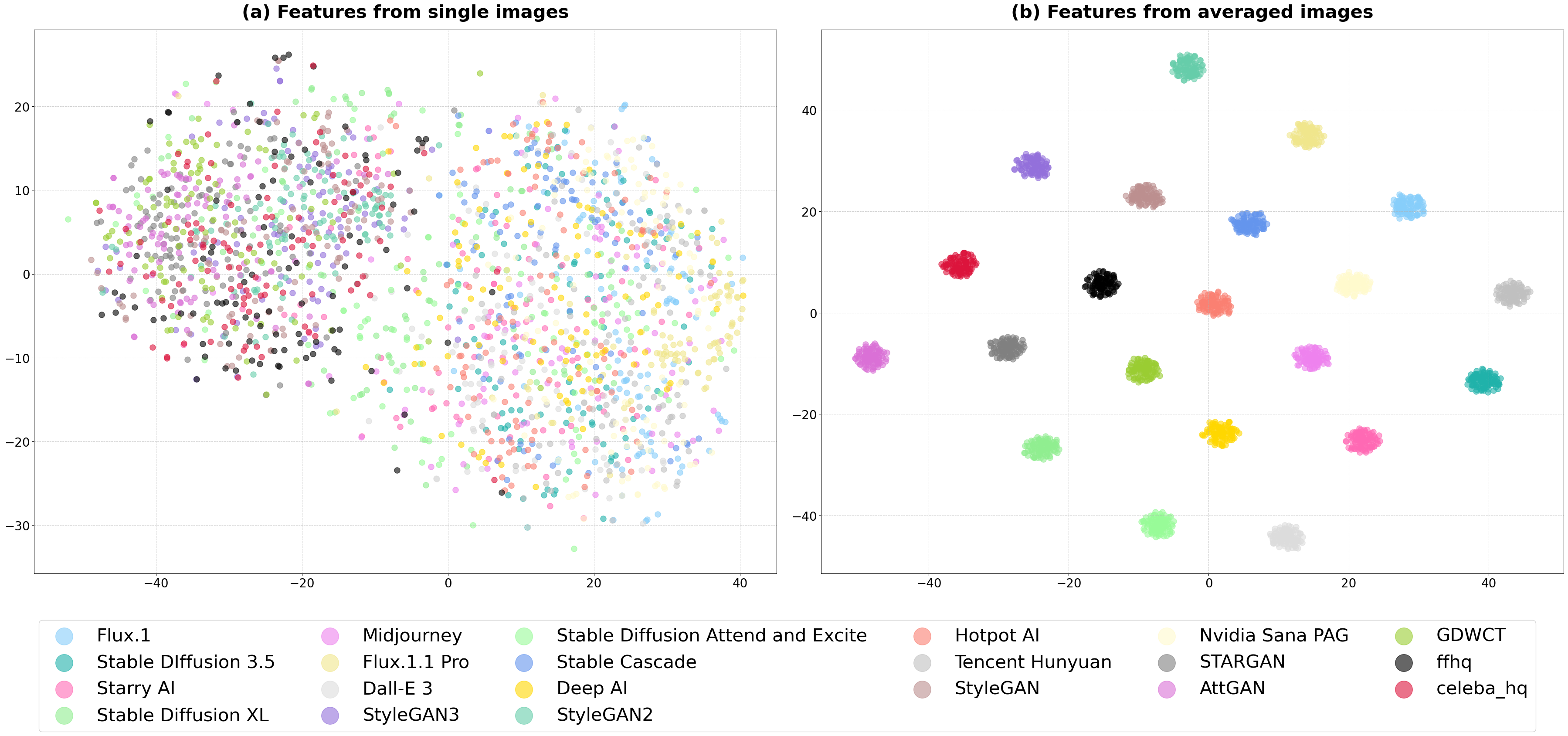}
    \caption{t-SNE analysis: (a) features extracted from single images; (b) features extracted from the average of images. The space of the average images exhibits a higher inter-class variability, yielding higher discriminative power for deepfake detection.}
    \label{fig:mean_features_sep}
\end{figure}

In this work, to improve the generalisation to unseen generators, we propose \textbf{\textit{\textit{\textmu}Flow}}, a novel deepfake face detector trained only on real faces without the need of generating pseudo-deepfakes or artificial artifacts. To detect deepfakes, we train \textbf{\textit{\textit{\textmu}Flow}} to map real samples in a pre-determined distribution and we treat log-likelihoods as \textit{fakeness} scores at inference time (\cref{fig:visual_abs}).
Previous methods \cite{khalid2020oc,CFLOW} project features into a  normal Gaussian distribution.
However, as shown in \cref{fig:mean_features_sep} (a), features extracted by real and fake samples exhibit low inter-class variability. Therefore, mapping features into a normal distribution results in high likelihoods for both classes, limiting current methods performance.

In contrast, we observe that averaging multiple faces significantly amplifies the distinctive traces injected by generative models \cite{burton,torralba,corvi2023detection}. As demonstrated in \cref{fig:mean_features_sep} (b), the representations extracted from the average images are clearly clustered among real and fake samples, making this feature space highly discriminative and suitable for deepfake detection.
However, despite their discriminative power, average images cannot be directly used at inference time, as we need to discriminate between individual real and fake images.
To overcome this issue, we propose leveraging the discriminative space derived from the average image by projecting into it the features extracted from single real images. The underlying intuition is that, once aligned with this discriminative space, features from individual images become inherently discriminative.

We benchmark \textbf{\textit{\textit{\textmu}Flow}} on three publicly available face datasets. Specifically, we evaluate its generalisation performance using a fully out-of-distribution evaluation setting. While trained solely on reals images from FFHQ~\cite{karras2019style}, it is tested on both unseen reals from CelebA-HQ~\cite{karras2018progressive} and deepfakes on unseen generators of the WILD~\cite{bongini2025wild} dataset. Under this severe generalisation setting, our method outperforms the state-of-the-art by a large margin.

\noindent The main contributions of our work are:
\begin{itemize}
    \item We approach deepfake face detection as a one-class classifier problem, training solely on real faces. Unlike previous methods, \textbf{\textit{\textit{\textmu}Flow}} does not require generating pseudo-deepfakes or synthetic artifacts, thus avoiding dependence on their quality and improving generalization.
    
    \item Based on the observation that average real image features are highly discriminative, we propose to align single-image features with such a space, enhancing their separability. We also provide an analytical derivation showing that mapping features into this discriminative space and maximizing their likelihood yields a direct criterion for distinguishing real from fake images.
    
    \item We validate our method on 19 unseen generators including both GANs and DMs architecture from the WILD dataset. \textbf{\textit{\textit{\textmu}Flow}} outperforms the SOTA by a large margin on several out-of-domain settings.
\end{itemize}

\section{Related Work}
\textbf{Deepfake Detection.} 
The detection of forged contents is an extremely active research field, aimed at analysing and detecting fake text, video, audio, as well as images \cite{Kaur2024}. Recently approaches have extensively focused on improving the generalisation of detectors, aiming at correctly identifying deepfakes from unseen generators.  FreqNet~\cite{tan2024frequency} analysed the high-frequencies  to find discriminative features. NPR~\cite{tan2024rethinking} analysed generator-specific traces using up-sampling layers. D$^3$ \cite{yang2025d} proposed to scale up the training set, including multiple generators to better learn such traces. Other approaches leveraged the generalisation capabilities of vision-language models such as CLIP, by prompt learning ~\cite{tan2025c2p} or finetuning \cite{CLIP2}. UFD \cite{UFD} was the first to show that a feature space not explicitly learned for generated image detection improve robustness due to its unbiased decision boundaries. Similarly, \cite{SVM} trained SVMs with the features from the penultimate layer, achieving good generalisation performance.
However, all of these approaches still train with both real and fake samples, inevitably limiting the generalisation to unseen generators.
Differently, we do not use any fake image during training, preventing the detector to  focus on specific traces.

\noindent \textbf{One-class Training for Deepfake Detectors.} 
To improve the generalisation, some methods proposed to train detectors using only real samples.
DiffFake \cite{diff_fake} use pairs of real faces from the same person to detect anomalous inconsistencies during inference. 
SeeABLE \cite{larue2023seeable} perturbs images to create pseudo-deepfakes, aligns them to prototypes, and detects anomalies by their distance from these prototypes.
UNTAG \cite{UNTAG} generated pseudo artifacts on real images and learns to detect them in a self-supervised manner.
CLIP-Flow \cite{yuan2025clip} combines CLIP features with a normalizing flow trained on real images and proxy samples. Similarly, \cite{shiohara2022detecting} synthesises self-blended images (SBI) from single real faces to reproduce general blending traces, while \cite{zhao2021learning} stitches real images via an inconsistency image generator (I2G) to produce pseudo-fakes. To capture local and semantic anomalies on pseudo-fakes, AUNet \cite{bai2023aunet} learns altered relations among synthetic facial movements, and \cite{nguyen2024laa} introduces a localized artifact attention network (LAA-Net) to detect fine-grained blending traces. Although trained only on real images, such methods still rely on synthesised pseudo-deepfakes or artifacts to obtain discriminative features, making their performance dependent on the quality of these generated proxies. OC-FakeDect \cite{khalid2020oc} did not generate artificial artifacts, as it trains a one-class Variational Autoencoder solely on real images and determines \textit{fakeness} scores via reconstruction errors. However, with the advent of novel generators, both real and fake images likely have similar features and thus similar reconstruction error, limiting their successful detection. 
Different from these approaches, \textbf{\textit{\textit{\textmu}Flow}} is trained solely on real faces and it does not require the generation of pseudo-deepfake samples or artifacts. Instead, it learns a discriminative features space by leveraging average images to better highlight traces injected by deepfake generators.

\noindent \textbf{Normalizing Flows.} 
Normalising flow models are designed to estimate complex data distributions using invertible transformations with tractable Jacobian determinants. By mapping data into a latent space, they enable exact likelihood computation \cite{glow}. FastFlow \cite{yu2021fastflow} and CFLOW-AD \cite{CFLOW} proposed to use normalizing flows for anomaly detection. Specifically, they train a normalizing to project image features into the standard normal distribution. During inference, likelihoods of transformed features are then treat as anomaly scores. Different from these methods, we map image features into an ad-hoc distribution where real and deepfakes can be easily discriminated.

\begin{figure*}[t!]
    \centering
    \includegraphics[width=1\linewidth]{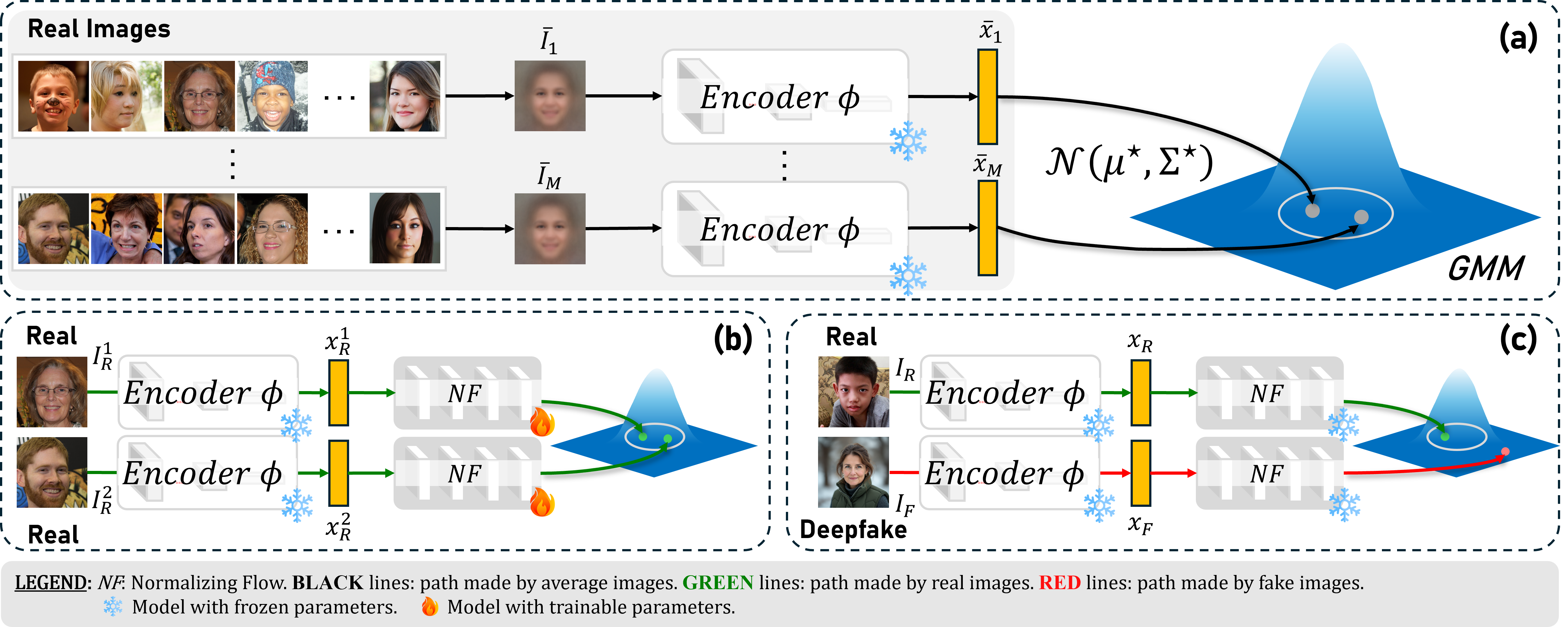}
    \caption{\textbf{\textit{\textmu}Flow} graphical overview. \textbf{(a) Discriminative Space Learning:} We learn a discriminative latent space extracting features from average real images (\cref{subsec:means}), using a Gaussian Mixture Model. \textbf{(b) FastFlow Training:} We train FastFlow to project representations extracted from real images into the learnt discriminative space (\cref{subsec: mapping}). \textbf{(c) Inference:} At test time, the representation of a real image is mapped in the region of the latent space with higher likelihood, while fake images are mapped to a low-likelihood regions (\cref{subsec:overall_fw}).}
    \label{fig:ad4dd}
\end{figure*}

\section{Proposed Method}

The workflow for \textbf{\textit{\textit{\textmu}Flow}} is depicted in \cref{fig:ad4dd}. We learn a discriminative latent space from the representation extracted by averaging sampled real faces  (\cref{fig:ad4dd}a). We train a normalizing flow \cite{yu2021fastflow} using features extracted from only real images that are mapped into this discriminative space (\cref{fig:ad4dd}b). As we trained with only real data, during inference, fake images obtain a lower likelihood and they are classified as deepfake.  (\cref{fig:ad4dd}c). In the sections below, we provide a detailed problem formulation and description of our method.

\subsection{Problem Formulation}
The task of detecting deepfakes can be formalised as follows: let $\mathcal{I} \subset \mathbb{R}^{H\times W\times3}$ be the image space and consider a dataset $\mathcal{D}=\mathcal{R}\cup\mathcal{F}$, composed of real faces $\mathcal{R}\subset\mathcal{I}$ and fake faces $\mathcal{F}\subset\mathcal{I}$. Fake samples are produced by a set of $N$ generators $\mathcal{G}=\{G_i\}_{i=1}^{N}$: for each $i$, let $\mathcal{F}^{(i)}=\{f^{(i)}_1,\ldots,f^{(i)}_{m_i}\}\subset\mathcal{I}$ denote the $m_i$ images generated by $G_i$, so that $\mathcal{F}\;=\;\bigcup_{i=1}^{N}\mathcal{F}^{(i)}.$ Our goal is to correctly distinguish between real and fake images for any input drawn from $\mathcal{D}$. 

Now, let split the $N$ generators into two disjoint sets for training and testing. The training set $\mathcal{D}_{\texttt{train}}=\mathcal{R}_{\texttt{train}}\cup(\textstyle\bigcup_{i=1}^{k}\mathcal{F}^{(i)})$ includes training real images $\mathcal{R}_{\texttt{train}}$ and deepfakes from $k<N$ generators, while the testing set $\mathcal{D}_{\texttt{test}}=\mathcal{R}_{\texttt{test}}\cup(\textstyle\bigcup_{i=k+1}^{N}\mathcal{F}^{(i)})$ is composed by testing real images $\mathcal{R}_{\texttt{test}}$ and deepfake from the other $N-k$ generators, with $\mathcal{R}=\mathcal{R}_{\texttt{train}}\cup\mathcal{R}_{\texttt{test}}$ and $ \mathcal{R}_{\texttt{train}} \cap \mathcal{R}_{\texttt{test}}=\emptyset$.

\noindent \textbf{Overview.} We train a one-class classifier to detect deepfakes, using solely real images, \textit{i.e., } \textbf{\textit{\textit{\textmu}Flow}} \textit{is not} provided with fake images during training, while still able to detect deepfakes from unseen generators. 
To this aim, all the generators are unseen to \textbf{\textit{\textit{\textmu}Flow}} and provided \textit{only} at testing time, by setting $k=0$,   $\mathcal{F}_{\texttt{train}} = \emptyset$ and $\mathcal{F}_{\texttt{test}} = (\textstyle\bigcup_{i=1}^{N}\mathcal{F}^{(i)})$.
This means that the training set for  \textbf{\textit{\textit{\textmu}Flow}} reduces to $\mathcal{D}_{\texttt{train}}\equiv \mathcal{R}_{\texttt{train}}$, and the testing set is $\mathcal{D}_{\texttt{test}}\equiv \mathcal{R}_{\texttt{test}}\cup\mathcal{F}_{\texttt{test}}$.

\subsection{Training a One-class Classifier for Deepfake Detection} 
Deepfake detection approaches are challenged in generalising to unseen generators~\cite{tan2024frequency, khan2023deepfake, yang2025d}. This is mainly because the detector learns to detect traces injected by training generators, which may differ from those injected by testing generators, especially if training and testing generators come from different categories (GANs vs. DMs)~\cite{corvi2023detection, Coccomini2023On}.

We train a one-class classifier that learns the distribution of real training samples, such that out-of-distribution samples are considered as fake. This allows our model to avoid overfitting on detecting traces injected by a specific generator, enhancing generalisability of our approach. Here, we leverage FastFlow~\cite{yu2021fastflow}, which is a state-of-the-art method for one-class detection.

\noindent\textbf{FastFlow.} During training, FastFlow extract features from images and trains a normalizing flow model to map all the extracted features from the raw distribution into a known distribution. %\hl{} 

Specifically, during training, an image $I_N \in \mathcal{I}$ is provided to a frozen encoder $\phi$, such that $x = \phi(I_N)$ represents the features for the sample $I_N$. 
A normalizing flow $f_{\theta}: X \rightarrow Z$, invertible with Jacobian $J_{f_\theta}(x)$, is used to project the image features $x \in p_X(x)$ into the hidden
variable $p_Z(z)$ with a bijective invertible mapping, where $p_X(x;\theta)\;=\;p_Z\!\big(f_\theta(x)\big)\,\big|\det J_{f_\theta}(x)\big|$. The prior density $p_Z=\mathcal{N}(0,I)$ is typically a Gaussian Normal distribution. The normalizing flow is trained by maximizing the expected log-likelihood of features from training samples, \textit{i.e.}, minimizing the negative log-likelihood (NLL):
\begin{equation}
    \begin{aligned}
    \mathcal{L}_{\mathrm{NLL}}(\theta)
    &= -\,\mathbb{E}_{\,x\sim X}\!\left[\log p_X\!\big(x\big)\right]
    = -\,\mathbb{E}_{\,x\sim P_X}\!\left[\log p_Z\!\big(z\big) + \log\big|\det J_{f_\theta}(x)\big|\right],
    \end{aligned}
\end{equation}
where $z=f_\theta(x)\sim p_Z$ and $x = f_\theta^{-1}(z)$. 

At testing time, samples that lie outside of the training distribution will have lower likelihoods and they will be detected and identified as outliers.

\subsection{Searching Generator Traces on Average Images} 
\label{subsec:means}

Following the one-class training paradigm, we train FastFlow~\cite{yu2021fastflow} only with real samples.
However, as shown in \cref{fig:mean_features_sep} (a), features extracted by a pretrained encoder from real and fake samples exhibit low inter-class variability. Consequently, FastFlow trained on such features tends to map both real and fake samples into the standard normal distribution, assigning high likelihoods to both classes and preventing effective separation.
State-of-the-art approaches overcome this issue by training the detectors on both real and fake images but losing generalisability on unseen generators. 

The current literature in deepfake detection has shown that averaging fake images amplifies the traces injected by generators \cite{Corvi2022OnTD}, making them easier to detect. As shown in \cref{fig:mean_features_sep} (b), features extracted from average real and fake images are clearly separable, showing a strong signal for deepfake detection.

\subsection{Projecting Features into the Average Image Feature Space} \label{subsec: mapping}
Although features extracted from average images are highly discriminative, they cannot be directly utilised during testing, as it is typically expected to perform inference with a single image.
\textit{How can we leverage the discriminative information encoded in the average images to effectively train a model capable of correctly classifying single images during inference?}

To this end, we propose to align the less discriminative feature space of single images with the highly discriminative feature space from average images. The underlying idea is that, by effectively mapping features from single samples into this discriminative space, these features become inherently discriminative. The next paragraphs provide an analytical derivation supporting this idea.
Following this intuition, building upon FastFlow, we train the normalizing flow using only real samples to project their features to the discriminative space from average real images. Different from Fastflow, our method projects the features to an ad-hoc discriminative distribution, rather than the standard normal distribution.

\noindent\textbf{Learning the underlying distribution of average real images.} 
To learn the mapping from \emph{single-image} real features to the discriminative space obtained from average images features, we first need to learn their underlying distribution.
To this aim, we generate a set of $M$ average real images $\{\overline{I}_R^j\}_{j=1}^{M}$ by averaging $K$ images randomly selected from the training real samples $\mathcal{R}_{train}$, as follows:

\begin{equation}
\label{eq:avg-feature}
\overline{I}_R^j \;:=\; \frac{1}{K}\sum_{i\in S_j} I_R^i,\qquad
\overline{x}_R^j \;:=\; \phi(\overline{I}_R^j),
\end{equation}
where $S_j$ is a set of randomly selected indices of  real training images and $\overline{x}_R^j$ are features extracted by the pretrained encoder $\phi: \mathcal{I} \rightarrow X$ on the average image $\overline{I}_R^j$.
We then collect these representations in the set $\mathcal{A}_{\mathcal{R}}=\{\overline{x}_R^1,\ldots,\overline{x}_R^M\}$ and we use this set to estimate the parameters $(\mu^\star,\Sigma^\star)$ by fitting a Gaussian mixture model (GMM) that encode the underlying distribution of average real features.

\noindent\textbf{Training with only real samples.} 
To leverage the discriminative power of features obtained from average images, we train the normalizing flow to map real features $x_R$ into the discriminative distribution $P_{Z}^* = \mathcal{N}(\mu^\star,\Sigma^\star)$ rather than a standard Gaussian normal distribution.

We hypothesise that real images are more correctly mapped on $P_{Z}^*$, compared to the fake images. This is equivalent to stating that the normalizing flow extracts latent representations from real images having a lower Mahalanobis distance with respect to our target distribution rather than fake images, thus providing a direct criterion to discriminate between real and fake images.

More formally, let $x_R = \phi(I_R) \in X$ be the representation of a real image $I_R$ and $z_R = f_\theta(x_R) \in Z$ be its latent vector extracted by the normalizing Flow $f_{\theta}: X \rightarrow Z$. The squared Mahalanobis distance $d_{P_Z^*}^2(z_R)$ of the latent vector $z_R$ to our target distribution $P_{Z}^*$ is defined as:

\begin{equation} \label{eq:maha}
d_{P_Z^*}^2(z_R) = (z_R - \mu^\star)^\top {\Sigma^\star}^{-1} (z_R - \mu^\star).
\end{equation}

\noindent For any randomly selected real $I_R$ and fake image $I_F$, we expect that:
\begin{equation} \label{eq:hypothesis}
d_{P_{Z}^*}^2(f_\theta(\phi(I_R))) < d_{P_{Z}^*}^2(f_\theta(\phi(I_F))).
\end{equation}

\noindent\textbf{Minimising the Mahalanobis distance for real images to the target distribution.} To enforce this hypothesis, we train the normalizing flow to maximise the log likelihoods for real image features $x_R$ from $P_Z^\star$, equivalently minimising NLL loss as follows:
\begin{equation} \label{eq:obj_nf}
\mathcal{L}_{\text{NLL}}(\theta) = - \mathbb{E}_{X_R \sim \mathcal{X}_{train}} \left[ \log p_X(x_R; \theta) \right],
\end{equation}
where $p_X(\cdot;\theta)$ is the probability density function (PDF) of the (unknown) distribution of real features $x_R$. Optimising this objective is equivalent to minimising the Mahalanobis distance of the latent representation from the real images $z_R$ to the target distribution $P_{Z}^*$. 

Indeed, $p_X(x_R; \theta)$ can be obtained as:
\begin{equation}
\label{eq:cov-ad4dd}
p_X(x_R;\theta)\;=\;p_{Z}^*\!\big(f_\theta(x_R)\big)\,\big|\det J_{f_\theta}(x_R)\big|.
\end{equation}

\noindent By using \cref{eq:cov-ad4dd}, we estimate the log-likelihood term in \cref{eq:obj_nf} as follows:

\begin{equation}
\label{eq:log_cov-ad4dd} 
\log p_X(x_R; \theta) = \log p_{Z}^*(z_R) +
\log\big|\det J_{f_\theta}(x_R) \big|.
\end{equation}

\noindent Given that $p_{Z}^*(z_R) = \mathcal{N}(z_R; \mu^\star, \Sigma^\star)$, we can expand the term $\log p_{Z}^*(z_R)$ as:

\begin{equation} \label{eq:log_s}
\log p_{Z}^*(z_R) = \log \left( \frac{\exp\left( -\frac{1}{2} ((z_R - \mu^\star)^\top
{\Sigma^\star}^{-1} 
(z_R - \mu^\star) \right)}{\sqrt{(2\pi)^D \det(\Sigma^\star)}} 
 \right),
\end{equation}

\noindent where $D$ is the dimension of the latent $z_R$. By substituting \cref{eq:maha} in \cref{eq:log_s}, we have:

\begin{equation}
\label{eq:not-sure-how-to-call-this}
\begin{aligned}
    \log p_{Z}^*(z_R) = \log \left( \frac{1}{\sqrt{(2\pi)^D \det(\Sigma^\star)}} \right) + \log \left( \exp\left( -\frac{1}{2} d^2_{P_{Z}^*}(z_R) \right) \right) \\
    = C - \frac{1}{2} d^2_{P_{Z}^*}(z_R),
\end{aligned}
\end{equation}

\noindent where $C = -\frac{1}{2} \left( D \log(2\pi) + \log \det(\Sigma^\star) \right)$ is a constant scalar term since $\Sigma^\star$ is precomputed. Hence, by integrating \cref{eq:not-sure-how-to-call-this} in \cref{eq:log_cov-ad4dd}, we obtain:

\begin{equation}
\label{eq:eq_9}
\log p_X(x_R; \theta) = C - \frac{1}{2} d^2_{P_{Z}^*}(z_R) +
\log\big|\det J_{f_\theta}(x_R) \big|.
\end{equation}

\noindent Plugging \cref{eq:eq_9} into \cref{eq:obj_nf}, we find that minimising \cref{eq:obj_nf} is equivalent to:
\begin{equation}
\underset{\theta}{\arg \min} \; - \mathbb{E}_{I_R \sim \mathcal{R}_{train}} \left[ C - \frac{1}{2} d^2_{P_{Z}^*}(f_{\theta}(\phi(I_R))) + \log \big| \det J_{f_\theta}(x_R) \big| \right],
\end{equation}
\noindent where $z_R = f_{\theta}(\phi(I_R))$.

\noindent Since $C$ is a constant, we can remove it from the minimisation objective:
\begin{equation}
\underset{\theta}{\arg \min} \;  \mathbb{E}_{I_R \sim \mathcal{R}_{train}} \left[ \frac{1}{2} d^2_{P_{Z}^*}(f_{\theta}(\phi(I_R))) - \log \big| \det J_{f_\theta}(x_R) \big| \right].
\end{equation}

\noindent The first term of this minimisation objective correspond to the first term of \cref{eq:hypothesis}. Consequently, this derivation explicitly shows that optimising \cref{eq:obj_nf} results in minimising $d^2_{P_{Z}^*}(z_R)$, e.g. the Mahalanobis distance of the real latent representations with respect to the discriminative distribution $P^*_{Z}$.
While our optimisation does not directly affect the second term in \cref{eq:hypothesis}, results in \cref{sec:exp} empirically show that minimising $d^2_{P_{Z}^*}(z_R)$ provides, at test time, a direct criterion to discriminate between real and fake latent representations.

\subsection{Overall Framework}
\label{subsec:overall_fw}

\cref{fig:ad4dd} shows the complete pipeline of \textbf{\textit{\textit{\textmu}Flow}}. Given a real image $I_R$, we first extract its feature representation $x_R = \phi(I_R)$ using a pretrained encoder $\phi:\mathcal{I}\rightarrow X$.
Building on the observation that features from averaged images are highly discriminative, we aim to model their underlying distribution.
To this end, we generate a set of averaged real images $\overline{I}_R$, extract their corresponding features $\overline{x}_R = \phi(\overline{I}_R)$, and fit a Gaussian distribution $P_Z^* = \mathcal{N}(\mu^\star, \Sigma^\star)$ that encode the discriminative power of this feature space.

\noindent \textbf{Training.} During training, a normalizing Flow $f_\theta:X \rightarrow Z$ is trained using only real samples to project single-image features $x_R$ into latent representation $z_R = f_{\theta}(x_R) \sim P_Z^*$ optimising the following NLL loss:

\begin{equation}
\underset{\theta}{\arg \min} \;  \mathbb{E}_{I_R \sim \mathcal{R}_{train}} \left[ \frac{1}{2} d^2_{P_{Z}^*}(z_R) - \log \big| \det J_{f_\theta}(x_R) \big| \right]
\end{equation}
Optimising this objective allows the model to learn a discriminative representation without relying on fake or pseudo-fake samples.

We then determine a threshold on collected likelihoods $L(\theta) = \frac{1}{2} d^2_{P_{Z}^*}(z_R) 
- \log \big| \det J_{f_\theta}(x_R) \big|$, as follows:

\begin{equation}
\label{eq:tau}
\tau = m + \gamma_{1-\alpha}\,\sigma,
\end{equation}

\noindent where $m = \operatorname{mean}(L(\theta))$ and $\sigma = \operatorname{std}(L(\theta))$ are the empirical mean and standard deviation of the training likelihoods, $\gamma_{1-\alpha} = \Phi^{-1}(1-\alpha)$, $\Phi$ is the cumulative distribution function of the normal distribution, and $\alpha \in (0,1)$ is the prescribed significance level. Under the Gaussian assumption that $L(\theta)$, $\alpha$ directly controls the theoretical false positive rate, \textit{i.e.}, $\mathbb{P}(L(\theta) > \tau) = \alpha$.

\noindent \textbf{Inference.} We compute the \textit{fakeness} score of a test image $I$ as:
\begin{equation}
s(I) = d^2_{P_{Z}^*}\big(f_\theta(\phi(I))\big) - \log \big| \det J_{f_\theta}(\phi(I)) \big|,
\end{equation}

\noindent where $\phi(I)$ are features extracted by the encoder $\phi$ and $f_\theta(\phi(I))$ are latent representations from the normalizing flow. 

We consider $I$ as fake if $s(I) > \tau$. A higher score indicates that the features lie far from the learned  distribution, as a result of detecting the distinctive traces injected by deepfake generators.

\section{Experimental Results}

\noindent\textbf{Datasets.} To ensure a fair evaluation, we adopted recent high-resolution human faces datasets: the \textbf{The Flickr‑Faces‑HQ (FFHQ)}~\cite{karras2019style} and \textbf{CelebA-HQ}~\cite{karras2018progressive} datasets for real images; and a subset of deepfake generators in the \textbf{WILD}~\cite{bongini2025wild} datasets used during testing. FFHQ dataset contains 70k images of human faces, offering significantly great diversity in age, ethnicity, and background. CelebA-HQ is a large-scale face attributes dataset with consisting of 30k aligned and cropped face images of celebrities. Both are widely used in generative modeling tasks due to its controlled structure and facial consistency across samples. WILD contains 20k high-resolution synthetic human faces, including text-to-image commercial generators and generators that simulate real-world “in-the-wild” conditions, \textit{i.e.}, obtained from both GANs and DMs and generated through both image-to-image and text-to-image approaches. To ensure a balanced representation of generative architecture, we added 3 additional GANs from \cite{guarnera2020deepfake}. Subsequently, we divided all the generators into the following categories:

\begin{itemize}
    \item \textbf{GANs} contains images synthesized by \textit{StyleGAN}~\cite{karras2019style}, \textit{StyleGAN2}~\cite{Karras2020StyleGAN2}, and \textit{StyleGAN3}~\cite{Karras2021AliasFree} from the WILD dataset, and supplemented by \textit{StarGAN}\cite{choi2018stargan}, \textit{GDWCT}\cite{cho2019image}, and \textit{AttGAN}\cite{he2019attgan} from~\cite{guarnera2020deepfake}.
    \item \textbf{Diffusion Models Open-source (DM-OS)} contains images generated by open-source DMs, including  \textit{Flux.1}~\cite{BFL2024Flux1}, \textit{Stable Diffusion 3.5}~\cite{StabilityAI2023SD3.5}, \textit{Stable Diffusion XL}~\cite{Podell2024SDXL}, \textit{Stable Cascade}~\cite{Lopez2024StableCascade}, \textit{Stable Diffusion Attend and Excite}~\cite{Chefer2023AttendExcite}.
    \item \textbf{Diffusion Models Closed-source (DM-CS)} contains images generated by closed-source DMs, including \textit{Dall-E 3}~\cite{OpenAI2024DALLE3}, \textit{Midjourney}~\cite{Midjourney2022}, \textit{Starry AI}~\cite{StarryAI2023}, \textit{Deep AI}~\cite{DeepAI2024}, \textit{Hotpot AI}~\cite{HotpotAI2024}, \textit{Nvidia Sana PAG}~\cite{xie2024sana}, \textit{Tencent Hunyuan}~\cite{li2024hunyuandit}, \textit{Flux.1.1 Pro}~\cite{BFL2024Flux1.1Pro}.
\end{itemize}

\noindent\textbf{Implementation details.} 
We used the ResNet50 as an encoder pretrained on ImageNet-1K~\cite{deng2009imagenet} with no fine-tuning. Supplementary material shows stable performance across different encoders. For the normalizing flow (NF), we use the same architecture as in~\cite{yu2021fastflow}. With the encoder frozen, the NF is trained on a single NVIDIA RTX A6000 with a batch size of 32 within 1000 epochs. We use the AdamW optimizer with an initial learning rate of $10^{-4}$ and a weight decay of $10^{-5}$. 
We set $\alpha=0.01$ in \cref{eq:tau}. In \cref{sec:analysis}, we show performance of our method with different values of $\alpha$.

\noindent\textbf{Baselines.}
We compare \textbf{\textit{\textit{\textmu}Flow}} with an extensive set of baselines including DFX-SN~\cite{pontorno2025deepfeaturex}, FreqNet~\cite{tan2024frequency}, NPR~\cite{tan2024rethinking}, ODDN~\cite{tao2024oddn}, and $D^3$~\cite{yang2025d}. 
Unlike \textbf{\textit{\textit{\textmu}Flow}}, these methods require both real and fake images in training. Consequently, we trained these models by iteratively using one of the three fake categories as fake training dataset, while assessing generalization at testing time on the remaining two. 
Furthermore, we include recent baselines that are also trained only on real images: AUNet\cite{bai2023aunet}, LAA-NET\cite{nguyen2024laa}, SBI\cite{shiohara2022detecting}, and PCL+I2G\cite{zhao2021learning}, OC-FakeDect\cite{khalid2020oc}, AFSD\cite{leyva2024data}. For fair comparisons, we trained all these baselines using their official implementations.

\begin{table}[t]
    \centering
    \caption{Comparisons of \textbf{\textit{\textit{\textmu}Flow}} with state-of-the-art approaches trained on different training sets (Real + GANs, Real + DM-CS, or Real + DM-OS), whereas our method is trained only on real images. \textsc{acc}: Accuracy. \textsc{auc}: Area Under the Curve. \textsc{ap}: Average Precision. DM-CS: Diffusion Model-Closed Source. DM-OS: Diffusion Model-Open Source. \textbf{Bold}: best performing method; \underline{underlined}: second-best method. \textit{Mix*} refers to detectors trained on fake samples generated with a subset of GANs and DMs, while still being tested on unseen architectures (GANs, DM-CS, and DM-OS).}
    \begin{adjustbox}{width=\textwidth}
    \renewcommand{\arraystretch}{0.9}
    \begin{tabular}{llc|ccc ccc ccc|ccc}
        \toprule
        \multicolumn{3}{l}{} & \multicolumn{9}{c}{\textbf{Out-of-domain Testing}} & \multicolumn{3}{c}{\textbf{Average}} \\ \cmidrule{1-15}     
        \multirow{2}{*}{Trained on} & \multirow{2}{*}{Method} & \multirow{2}{*}{Venue} & \multicolumn{3}{c}{GANs} & \multicolumn{3}{c}{DM-CS} & \multicolumn{3}{c|}{DM-OS} & & \\
        &  &  & \textsc{acc} $\uparrow$ & \textsc{auc} $\uparrow$& \textsc{ap} $\uparrow$& \textsc{acc} $\uparrow$& \textsc{auc} $\uparrow$& \textsc{ap} $\uparrow$& \textsc{acc} $\uparrow$& \textsc{auc} $\uparrow$& \textsc{ap} $\uparrow$& \textsc{acc} $\uparrow$& \textsc{auc} $\uparrow$& \textsc{ap} $\uparrow$\\ \midrule \midrule
        
        \multirow{5}{*}{\shortstack{Real\\+\\GANs}}
        & DFX-SN~\cite{pontorno2025deepfeaturex}    & MTA25  & - & - & - & 74.1 & 71.7 & 69.9 & 72.5 & 69.0 & 67.0 & 73.3 & 70.1 & 68.0 \\
        & FreqNet~\cite{tan2024frequency}           & AAAI24 & - & - & - & 76.4 & 75.5 & 74.1 & 77.0 & 73.2 & 74.2 & 76.7 & 74.3 & 75.4 \\
        & NPR~\cite{tan2024rethinking}              & CVPR24 & - & - & - & 78.6 & 79.2 & 78.9 & 76.1 & 77.3 & 80.1 & 77.4 & 78.2 & 79.7 \\
        & ODDN~\cite{tao2024oddn}                   & AAAI25 & - & - & - & \underline{81.4} & \underline{83.2} & 84.8 & \underline{82.2} & 82.9 & 85.3 & \underline{81.8} & \underline{83.6} & 85.0 \\
        & D$^3$~\cite{yang2025d}                    & CVPR25 & - & - & - & 80.5 & 82.0 & \underline{88.8} & 81.3 & \underline{83.1} & \underline{87.}0 & 80.9 & 82.0 & \underline{87.9} \\ 
        \rowcolor{mygray} 
        Real-only& \textbf{\textit{\textit{\textmu}Flow} (Ours)}                         &  & - & - & - & \textbf{96.8} & \textbf{96.8} & \textbf{95.9} & \textbf{96.9} & \textbf{96.8} & \textbf{96.1} & \textbf{96.8} & \textbf{96.8} & \textbf{96.0} \\ \midrule \midrule
        
        \multirow{5}{*}{\shortstack{Real\\+\\DM-CS}}
        & DFX-SN~\cite{pontorno2025deepfeaturex}    & MTA25  & 67.9 & 67.0 & 66.1 & - & - & - & 80.2 & 84.6 & 86.0 & 74.1 & 75.8 & 76.0 \\
        & FreqNet~\cite{tan2024frequency}           & AAAI24 & 68.3 & 66.0 & 65.3 & - & - & - & 86.9 & 94.0 & 92.8 & 77.6 & 80.0 & 79.0 \\
        & NPR~\cite{tan2024rethinking}              & CVPR24 & 68.5 & 68.5 & 69.1 & - & - & - & 85.9 & 93.4 & 92.7 & 77.2 & 81.0 & 80.9 \\
        & ODDN~\cite{tao2024oddn}                   & AAAI25 & \underline{84.5} & 79.2 & 75.9 & - & - & - & \underline{97.5} & \textbf{98.5} & 91.7 & \underline{91.0} & 88.8 & 83.8 \\
        & D$^3$~\cite{yang2025d}                    & CVPR25 & 75.1 & \underline{83.4} & \underline{85.0} & - & - & - & \textbf{98.6} & \underline{97.9} & \textbf{99.9} & 86.8 & \underline{90.7} & \underline{92.5} \\
        \rowcolor{mygray} 
        Real-only& \textbf{\textit{\textit{\textmu}Flow} (Ours)}                         &  & \textbf{90.3} & \textbf{90.4} & \textbf{93.9} & - & - & - & 96.9 & 96.8 & \underline{96.1} & \textbf{93.6} & \textbf{93.6} & \textbf{95.0} \\ \midrule \midrule
        
        \multirow{5}{*}{\shortstack{Real\\+\\DM-OS}}
        & DFX-SN~\cite{pontorno2025deepfeaturex}    & MTA25  & 69.9 & 70.4 & 68.6 & 81.2 & 80.7 & 83.0 & - & - & - & 75.6 & 75.6 & 75.8 \\
        & FreqNet~\cite{tan2024frequency}           & AAAI24 & 71.8 & 67.6 & 70.7 & 87.7 & 94.7 & 94.4 & - & - & - & 79.8 & 81.2 & 82.6 \\
        & NPR~\cite{tan2024rethinking}              & CVPR24 & 75.1 & 68.7 & 73.5 & 92.1 & 97.7 & \textbf{98.0} & - & - & - & 83.6 & 83.2 & 85.8 \\
        & ODDN~\cite{tao2024oddn}                   & AAAI25 & \underline{85.2} & 71.2 & 70.3 & \underline{98.4} & \textbf{99.5} & 96.7 & - & - & - & \underline{91.8} & 85.3 & 83.5 \\
        & D$^3$~\cite{yang2025d}                    & CVPR25 & 71.1 & \underline{76.6} & \underline{77.9} & \textbf{99.7} & \underline{98.0} & \underline{97.1} & - & - & - & 85.4 & \underline{87.3} & \underline{87.5} \\ 
        \rowcolor{mygray}
        \rowcolor{mygray}
        Real-only& \textbf{\textit{\textit{\textmu}Flow} (Ours)}                         &  & \textbf{90.3} & \textbf{90.4} & \textbf{93.9} & 96.8 & 96.8 & 95.9 & - & - & - & \textbf{93.5} & \textbf{93.6} & \textbf{94.9} \\ \midrule \midrule
        
        \multirow{5}{*}{\shortstack{Real\\+\\Mix$^*$}}
        & DFX-SN~\cite{pontorno2025deepfeaturex}    & MTA25  & 75.2 & 77.9 & 77.1 & 77.0 & 80.4 & 82.2 & 81.2 & 81.9 & 82.0 & 77.8 & 80.1 & 80.4 \\
        & FreqNet~\cite{tan2024frequency}           & AAAI24 & 79.9 & 86.0 & 87.5 & 77.9 & 84.6 & 84.6 & 73.5 & 73.2 & 76.8 & 77.1 & 81.3 & 83.0 \\
        & NPR~\cite{tan2024rethinking}              & CVPR24 & 83.0 & 89.8 & 89.5 & 87.4 & \underline{96.5} & 90.2 & 85.4 & 95.7 & 95.4 & 85.3 & 94.0 & 91.7 \\
        & ODDN~\cite{tao2024oddn}                   & AAAI25 & \textbf{92.7} & \underline{91.5} & 93.5 & 90.3 & 83.3 & 86.5 & 91.1 & 86.3 & 85.6 & 91.4 & 87.0 & 88.5 \\
        & D$^3$~\cite{yang2025d}                    & CVPR25 & \underline{90.4} & \textbf{92.5} & \underline{93.7} & \underline{95.8} & 94.6 & \underline{92.6} & \underline{91.7} & \underline{96.5} & \textbf{96.4} & \underline{92.6} & \underline{94.5} & \underline{94.2} \\
        \rowcolor{mygray}
        Real-only& \textbf{\textit{\textit{\textmu}Flow} (Ours)}                       &  & 90.3 & 90.4 & \textbf{93.9} & \textbf{96.8} & \textbf{96.8} & \textbf{95.9} & \textbf{96.9} & \textbf{96.8} & \underline{96.1} & \textbf{94.7} & \textbf{94.7} & \textbf{95.3} \\
        \midrule \midrule     
        
        \multirow{7}{*}{\shortstack{Real-only}}
        & OC-FakeDect~\cite{khalid2020oc} & CVPR20 & 48.9 & 51.1 & 50.0 & 62.9 & 62.3 & 59.8 & 60.4 & 60.9 & 59.8 & 57.4 & 58.1 & 56.5 \\
        & AFSD~\cite{leyva2024data} & PRL & 67.3 & 69.0 & 69.3 & 64.7 & 61.3 & 67.0 & 63.0 & 62.9 & 70.1 & 65.0 & 64.4 & 68.8 \\
        & PCL+I2G\cite{zhao2021learning} & ICCV21 & 84.1 & 86.0 & 92.2 & 75.0 & 78.7 & 69.3& 73.6 & 73.0 & 68.3 & 77.6 & 79.2 & 76.6 \\
        & SBI\cite{shiohara2022detecting}& CVPR22 & 84.3 & 86.8 & 93.0 & 75.2 & 79.4 & 69.7 & 74.0 & 73.2 & 68.5 & 77.8 & 79.8 & 77.1 \\
        & AUNet\cite{bai2023aunet}   & CVPR23 & 84.8& 87.0& 93.3& 75.3& 80.0& 70.2& 74.3& 73.5& 70.1& 78.1 & 80.2 & 77.9 \\
        & LAA-Net\cite{nguyen2024laa}& CVPR24 & \underline{86.0}& \underline{88.2}& \textbf{94.2}& \underline{83.2}& \underline{84.1}& \underline{81.0}& \underline{79.9}& \underline{82.3}& \underline{83.3} & \underline{83.0} & \underline{84.9} & \underline{86.2} \\ %\cmidrule{2-15}
        \rowcolor{mygray}
        & \textbf{\textit{\textit{\textmu}Flow} (Ours)}                       &  & \textbf{90.3} & \textbf{90.4} & \underline{93.9} & \textbf{96.8} & \textbf{96.8} & \textbf{95.9} & \textbf{96.9} & \textbf{96.8} & \textbf{96.1} & \textbf{94.7} & \textbf{94.7} & \textbf{95.3} \\ \bottomrule
    \end{tabular}
    \end{adjustbox}
    \label{tab:main}
\end{table}

\subsection{Results} \label{sec:exp}
We evaluate the generalisation capability of \textbf{\textit{\textit{\textmu}Flow}} by training it on real images from FFHQ \cite{karras2019style}, while testing on the unseen fakes from WILD \cite{bongini2025wild}. Note that, at testing time, we use unseen reals from CelebA-HQ \cite{karras2018progressive} to further assess the ability of \textbf{\textit{\textit{\textmu}Flow}} to generalise across both unseen reals and fake images. 

\noindent\textbf{Generalisation results across generator categories.} \cref{tab:main} {shows the OOD generalization performance on unseen generators and compares our method with several recent SOTA baselines trained on both real and fake images.} We group results based on training dataset used to train SOTA methods: \textit{Real+GANs}, \textit{Real+DM-CS}, and \textit{Real+DM-OS}. Specifically, methods trained with \textit{Real+GANs} use real images together with deepfakes generated by GANs.
Methods in \textit{Real+DM-CS} are trained with real images and deepfakes generated using closed-source diffusion model implementations, while \textit{Real+DM-OS} methods use deepfakes produced by open-source diffusion models.
Note that for all the configurations, our method is trained \textit{only} with real samples.

In \textit{Real+GANs}, all compared methods underperform into generalising to diffusion-based forgeries (DM-CS and DM-OS). The average \textsc{acc} across the three OOD test domains is significantly lower than ours. The second best-performing comparison method in this setting, ODDN, achieves an average \textsc{acc} of $81.8$, whereas our method reaches $96.8$, yielding a significant improvement of $+15$ in \textsc{acc}, respectively. These results show that SOTA methods struggle in generalising to different categories of generators (GANs vs. DMs). In the \textit{Real+DM-CS} training setting, the performance improves substantially, particularly for ODDN. By training on DM-CS, $D^3$ increases performance on other DMs (DM-OS), showing its ability to generalise within the same category of generators (DMs). Nonetheless, our real-only approach still outperforms it ($+2.6$ \textsc{acc}). Similarly, in the \textit{Real+DM-OS} scenario we observe the strongest performance among the SOTA methods, particularly with $D^3$ ($85.4$ \textsc{acc}). However, our method still outperforms the SOTA ($+1.7$ \textsc{acc}).

\begin{table}[t]
    \centering
    \caption{Performance under several content-preserving transformations.}
    \renewcommand{\arraystretch}{1.1}
    \begin{adjustbox}{width=\columnwidth}
    \begin{tabular}{l ccc |ccc |ccc |ccc||ccc}
    \toprule
    \multirow{2}{*}{Method} & 
    \multicolumn{3}{c}{Resize} & 
    \multicolumn{3}{c}{Horizontal Flip} & 
    \multicolumn{3}{c}{Gaussian Noise} &
    \multicolumn{3}{c}{Salt Pepper} & \multicolumn{3}{c}{\textbf{Average}}\\
    \cmidrule{2-16}
     & \textsc{acc} $\uparrow$ & \textsc{auc} $\uparrow$ & \textsc{ap} $\uparrow$ & \textsc{acc} $\uparrow$ & \textsc{auc} $\uparrow$ & \textsc{ap} $\uparrow$ & \textsc{acc} $\uparrow$ & \textsc{auc} $\uparrow$ & \textsc{ap} $\uparrow$ & \textsc{acc} $\uparrow$ & \textsc{auc} $\uparrow$ & \textsc{ap} $\uparrow$ & \textsc{acc} $\uparrow$ & \textsc{auc} $\uparrow$ & \textsc{ap} $\uparrow$\\
    \midrule
DFX-SN~\cite{pontorno2025deepfeaturex}  & 72.1 & 73.0 & 73.5 & 71.8 & 75.3 & 74.1 & 70.9 & 74.2 & 72.4 & 74.9 & 72.9 & 76.4 & 72.4 & 73.8 & 74.1 \\
    FreqNet~\cite{tan2024frequency}         & 76.0 & 81.8 & 81.7 & 76.3 & 74.3 & 79.9 & 75.0 & 78.8 & 76.4 & 73.9 & 79.8 & 77.3 & 75.3 & 78.7 & 78.8 \\
    NPR~\cite{tan2024rethinking}            & 85.0 & \underline{92.8} & 91.1 & 84.4 & 87.9 & 88.0 & 77.2 & 80.7 & 79.9 & 70.3 & 74.1 & 75.2 & 79.2 & 83.9 & 83.5 \\
    ODDN~\cite{tao2024oddn}                 & \underline{89.4} & \textbf{94.8} & \textbf{95.9} & 89.3 & 90.2 & \textbf{92.7} & 76.9 & 74.5 & 74.9 & 75.2 & 77.7 & 76.1 & 82.7 & 84.3 & 84.9 \\
    $D^3$~\cite{yang2025d}                  & 89.1 & 91.7 & \underline{91.3} & \underline{90.3} & \underline{91.8} & 91.3 & \underline{88.2} & \textbf{91.9} & \textbf{92.0} & \underline{90.0} & \underline{90.1} & \textbf{93.8} & \underline{89.4} & \underline{91.4} & \textbf{92.1} \\ 
    PCL+I2G\cite{zhao2021learning}          & 76.0 & 76.1 & 78.7 & 75.9 & 76.5 & 79.8 & 68.8 & 74.7 & 77.2 & 65.5 & 74.7 & 76.2 & 71.5 & 75.5 & 78.0 \\
    SBI\cite{shiohara2022detecting}         & 76.4 & 76.3 & 80.1 & 77.0 & 77.1 & 80.5 & 69.9 & 75.1 & 77.9 & 66.1 & 75.3 & 76.9 & 72.3 & 75.9 & 78.8 \\
    AUNet\cite{bai2023aunet}                & 76.6 & 76.7 & 80.0 & 77.4 & 77.4 & 80.2 & 70.3 & 75.8 & 78.4 & 67.0 & 75.8 & 77.5 & 72.8 & 76.4 & 79.0 \\
    LAA-NET\cite{nguyen2024laa}             & 82.3 & 83.0 & 84.7 & 81.3 & 82.5 & 83.4 & 81.0 & 82.1 & 80.9 & 79.9 & 78.7 & 80.1 & 81.1 & 81.6 & 82.3 \\ \midrule \rowcolor{mygray}
    \textbf{\textit{\textit{\textmu}Flow} (Ours)}                           & \textbf{90.6} & 91.5 & 88.4 & \textbf{92.6} & \textbf{92.7} & \underline{91.9} & \textbf{89.7} & \underline{89.9} & \underline{91.7} & \textbf{90.8} & \textbf{91.7} & \underline{89.6} & \textbf{90.9} & \textbf{91.5} & \underline{90.4} \\
    \bottomrule
    \end{tabular}
    \end{adjustbox}
    \label{tab:robustness}
\end{table}

\noindent\textbf{Generalisation results mixing generator categories.} Here, we do not separate deepfakes by generator category. All competing methods are trained on a mixed dataset containing both GANs and DMs images, while testing on unseen architectures from both categories.
As shown in~\cref{tab:main} (\textit{Real+Mix$^*$} group) competing methods benefit from training from a heterogeneous set of generators. This strengthens our hypothesis that such methods learn specific traces injected by generators and thus benefit of seeing multiple generators during training. $D^3$ and ODDN perform strongly, reaching \textsc{acc} of $92.6$ and $91.4$, respectively. However, despite our method is trained only on real images, \textbf{\textit{\textit{\textmu}Flow}} obtains an average \textsc{acc} $94.7$ \textsc{acc}, improving the second-best performing approach by $+2.1$.

\noindent\textbf{Comparisons with real-only training baselines.}
Finally, we compare \textbf{\textit{\textit{\textmu}Flow}} with prior real-only training approaches, namely PCL+I2G, SBI, AUNet, and LAA-Net. While these methods do not rely on fake samples during training, their generalisation capability remains limited. In particular, the best-performing baseline in this group, LAA-Net, achieves an average \textsc{acc} of $83.0$, whereas our method reaches $94.7$, yielding a substantial improvement of $+11.7$. The gap is especially pronounced on diffusion-based forgeries (DM-CS and DM-OS), where our approach consistently obtains nearly $97\%$ \textsc{acc}, largely outperforming all competitors. These results demonstrate that simply training on real images is not sufficient to ensure strong OOD generalisation, which is instead achieved by leveraging the discriminative power of average images.

\begin{figure}[t]
    \centering
    \includegraphics[width=\linewidth]{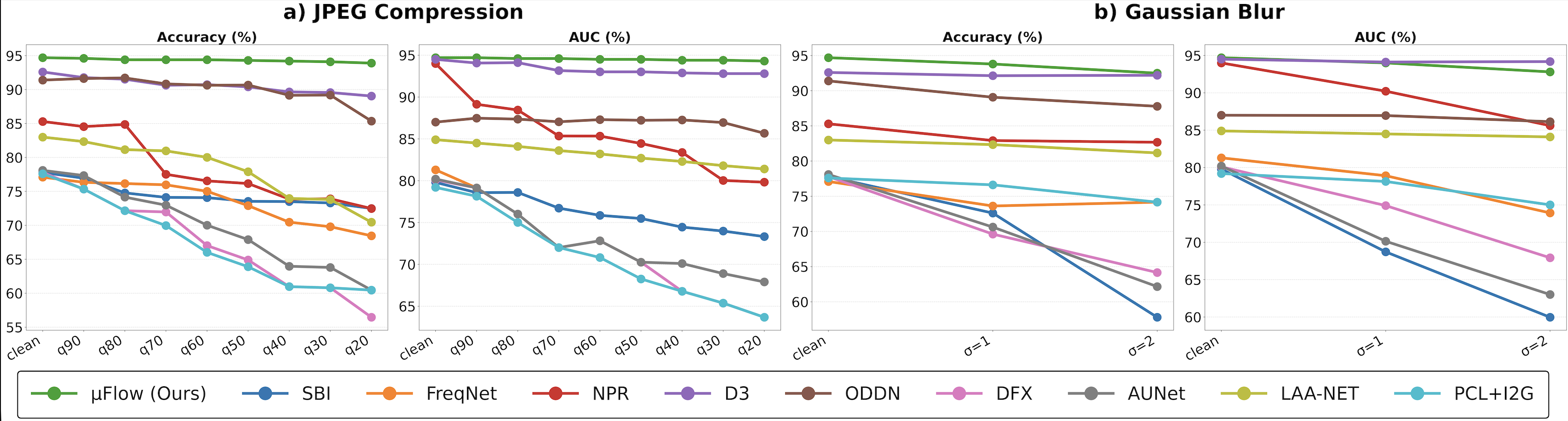}
    \caption{\textsc{acc} and \textsc{auc} under inference-time degradations: (a) JPEG compression with decreasing quality factor; (b) Gaussian blur with increasing standard deviation.}
    \label{fig:robustness}
\end{figure}

\subsection{Analysis}
\label{sec:analysis}

\noindent\textbf{Robustness Analysis.} \cref{tab:robustness} shows the robustness of our approach under content-preserving transformations (resize, horizontal flip, Gaussian noise, salt pepper). Although \textbf{\textit{\textit{\textmu}Flow}} is not trained against these transformations, performance marginally reduces, achieving comparable results with respect to the state-of-the-art. \cref{fig:robustness} shows experiments by applying signal-degrading perturbations to test images. Overall, \textbf{\textit{\textit{\textmu}Flow}} is not highly impacted by such perturbations, compared to the state-of-the-art. Moreover, it maintains stable performance with the increasing of perturbation intensity, showing the robustness of our latent representations.

\begin{figure}[t!]
    \centering
    
    \begin{minipage}{0.5\linewidth}
        \captionof{table}{Ablation study on \textbf{\textit{\textit{\textmu}Flow}} across target distributions and inference classifiers.}
        \centering
        \begin{adjustbox}{max width=0.8\linewidth}
        \begin{tabular}{ccc|c}
        \toprule
            \textbf{\makecell{Target  distribution}} & \textbf{Classifier} & \textbf{$\alpha$} & \textsc{acc}\\
        \midrule
            \multirow{5}{*}{$\mathcal{N}(0, \mathbb{I})$} & LOF & - & 59.1\\
            & OC-SVM & - & 62.1\\
            & \multirow{3}{*}{$\tau$} & $0.1$ & 54.8\\
            && $0.05$ & 57.6\\
            && $0.01$ & 59.9\\
            \midrule
            \multirow{5}{*}{$\mathcal{N}(\mu^\star, \Sigma^\star)$} & LOF & - & \underline{94.2}\\
             & OC-SVM & - & 93.9\\
             & \multirow{3}{*}{$\tau$} & $0.1$ & 89.7\\
            && $0.05$ & 93.9\\
            && $0.01$ & \textbf{94.7}\\
            \bottomrule
        \end{tabular}
        \end{adjustbox}
        \label{tab:ablation}
    \end{minipage}
    \hfill
    \begin{minipage}{0.49\linewidth}
        \centering
        \includegraphics[width=\linewidth]{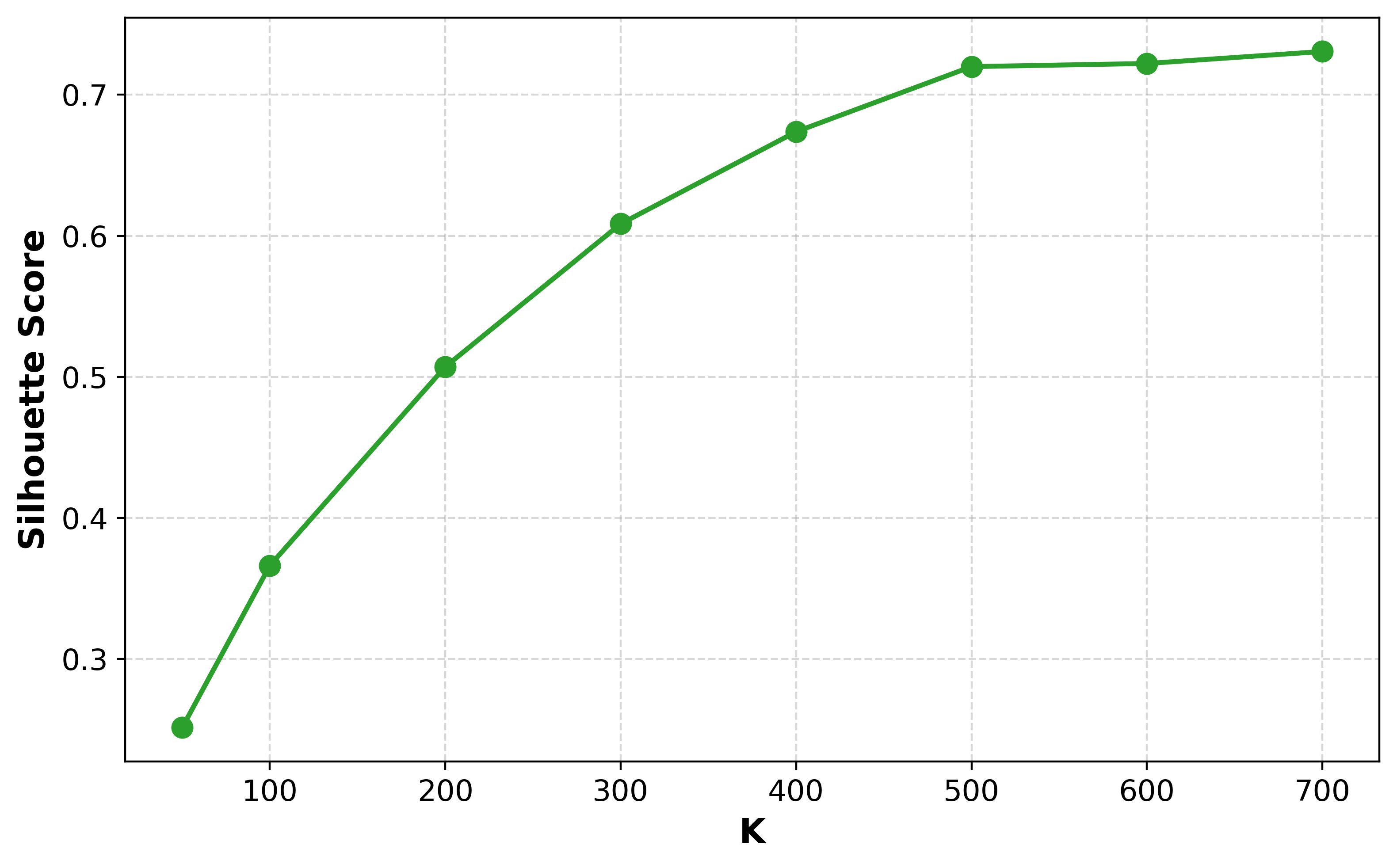}
        \captionof{figure}{Silhouette score with different values of $K$.}
        \label{fig:clusters}
    \end{minipage}
\end{figure}

\noindent\textbf{Ablation Studies.} In this section, we analyse contributions of each component of our method. Specifically, we compare results using the discriminative distribution $\mathcal{N}(\mu^*\Sigma^*)$ rather than the normal Gaussian distribution. Moreover, we compare the use of the fixed threshold $\tau$ wrt. one-class naive machine learning models, such as Local Outlier Factor (LOF) and One-class Support Vector Machine (OC-SVM). \cref{tab:ablation} (top part) show that using the normal Gaussian distribution fails to separate the classes. Differently, using the discriminative space $\mathcal{N}(\mu^*\Sigma^*)$ learnt from average images, our method drastically increase performance, validating our hypothesis to leverage the discriminative power of average images. \cref{tab:ablation} (bottom part) shows that our method obtains similar performances using both a fixed threshold or naive machine learning models.
Finally, we report an ablation study over different values of $\alpha$. 
Although the overall performance remains stable across the explored range, we observe a consistent—albeit moderate—improvement for $\alpha = 0.01$, which yields the best trade-off between detection sensitivity and false positive control. Consequently, we adopt $\alpha = 0.01$ in all subsequent experiments.

\noindent\textbf{Evaluating sample size for average images.} 
The training of \textbf{\textit{\textit{\textmu}Flow}} also relies on the choice of the number $K$ of training images to use to determine the average images (\cref{eq:avg-feature}). Intuitively, larger $K$ better suppresses image-specific noise while amplifying generator-specific traces, yielding more discriminative features. 
To choose the best value of $K$, we compute the \textit{Silhouette Score} across a wide range of values (\cref{fig:clusters}). This metric evaluates how much clusters are separated and thus the discriminative power of features. The score follows a logarithmic growth trend, reaching $0.72$ at $K=500$ and $0.74$ at $K=700$. Although average images are computed only once prior to training, higher values of $K$ imply an increasing computational time. We therefore set $K=500$ as a trade-off between discriminative power and efficiency.

\section{Conclusion \& Limitations}

In this paper, we introduced \textbf{\textit{\textit{\textmu}Flow}}, an out-of-distribution deepfake detector. Unlike the state-of-the-art, we train our model with real images only and we use a modified formulation of normalised flow to take advantage of the discriminative power of average images. Results on real and generated images of faces have shown that \textbf{\textit{\textit{\textmu}Flow}} outperforms the state-of-the-art across the board. Nonentheless, our work presents some limitations, being restricted to detecting face deepfakes. As future work, we plan to extend the method to handle multiple semantic categories, as well as apply it to videos.

\section*{Acknowledgements} Orazio Pontorno is a PhD candidate enrolled in the National PhD in Artificial Intelligence, XXXIX cycle, organized by Università Campus Bio-Medico di Roma. \\ 
This work was supported by the DEFORM project, funded by the European Union's Horizon Europe Research and Innovation Programme under Grant Agreement No. 101308502.

% ---- Bibliography ----
%
% BibTeX users should specify bibliography style 'splncs04'.
% References will then be sorted and formatted in the correct style.
%
\bibliographystyle{splncs04}
\bibliography{main}
\end{document}